\documentclass{article} 

\usepackage{nips15submit_e,times}
\usepackage{hyperref}
\usepackage{url}
\usepackage{times}
\usepackage{graphicx} 
\usepackage{subfig}
\usepackage[numbers]{natbib}
\usepackage{algorithmicx}
\usepackage{algpseudocode}
\usepackage{algorithm}
\usepackage{amsmath}
\usepackage{amssymb}
\usepackage{hyperref}
\usepackage[font=small,skip=5pt]{caption}
\newcommand{\matvec}{\mathbf{vec}}

\title{Natural Neural Networks}

\author{
Guillaume Desjardins, Karen Simonyan, Razvan Pascanu, Koray Kavukcuoglu\\
\texttt{\{gdesjardins,simonyan,razp,korayk\}@google.com} \\
Google DeepMind, London \\
}

\nipsfinalcopy 

\begin{document}

\maketitle

\begin{abstract}
We introduce Natural Neural Networks, a novel family of algorithms that speed up convergence by adapting their internal representation during training to improve conditioning of the Fisher matrix. In particular, we show a specific example that employs a simple and efficient reparametrization of the neural network weights by implicitly whitening the representation obtained at each layer, while preserving the feed-forward computation of the network. Such networks can be trained efficiently via the proposed Projected Natural Gradient Descent algorithm (PRONG), which amortizes the cost of these reparametrizations over many parameter updates and is closely related to the Mirror Descent online learning algorithm. We highlight the benefits of our method on both unsupervised and supervised learning tasks, and showcase its scalability by training on the large-scale ImageNet Challenge dataset.
\end{abstract}

\section{Introduction}
\label{intro}

Deep networks have proven extremely successful across a broad range
of applications. While their deep and complex structure affords them a rich
modeling capacity, it also creates complex dependencies between the parameters
which can make learning difficult via first order stochastic gradient descent
(SGD). As long as SGD remains the workhorse of deep learning, our ability to
extract high-level representations from data may be hindered by difficult
optimization, as evidenced by the boost in performance offered by batch
normalization (BN) \cite{batchnorm} on the Inception architecture
\cite{inception}.

Though its adoption remains limited, the natural gradient \cite{amari98natural}
appears ideally suited to these difficult optimization issues. By
following the direction of steepest descent on the probabilistic manifold, the
natural gradient can make constant progress over the course of optimization, as
measured by the Kullback-Leibler (KL) divergence between consecutive iterates.
Utilizing the proper distance measure ensures that the natural gradient is
invariant to the parametrization of the model. Unfortunately, its application
has been limited due to its high computational cost. Natural
gradient descent (NGD) typically requires an estimate of the Fisher Information
Matrix (FIM) which is square in the number of parameters, and worse, it
requires computing its inverse.  Truncated Newton methods can avoid
explicitly forming the FIM in memory \cite{martens2010hessian,
Pascanu-natural-arxiv2013}, but they require an expensive iterative procedure
to compute the inverse.
Such computations can be wasteful and do not take into account the smooth change of the  Fisher during optimization
or the highly structured nature of deep models.

Inspired by recent work on model reparametrizations \cite{Raiko-2012-small,
Montavon2013-centering}, our approach starts with a simple question: can we
devise a neural network architecture whose Fisher is constrained to be
identity? This is an important question, as SGD and NGD would be equivalent
in the resulting model.
The main contribution of this paper is in providing a simple, theoretically
justified network reparametrization which approximates via first-order
gradient descent, a block-diagonal
natural gradient update over layers.
Our method is computationally efficient due to the local nature of the
reparametrization, based on whitening, and the amortized nature of the
algorithm.
Our second contribution is in unifying many heuristics commonly used
for training neural networks, under the roof of the natural gradient, while
highlighting an important connection between model reparametrizations
and Mirror Descent \cite{Beck03}.
Finally, we showcase the efficiency and the scalability of our method across a
broad-range of experiments, scaling our method from standard deep auto-encoders
to large convolutional models on ImageNet\cite{ILSVRC15}, trained across multiple
GPUs.  This is to our knowledge the first-time a (non-diagonal) natural
gradient algorithm is scaled to problems of
this magnitude.

\section{The Natural Gradient}
\label{sec:background}

This section provides the necessary background and derives a particular form
of the FIM
whose structure will be key to our
efficient approximation. While we tailor the development of our method to the
classification setting, our approach generalizes to regression and density
estimation.

\subsection{Overview}

We consider the problem of fitting the parameters $\theta \in \mathbb{R}^N$ of a
model $p(y \mid x; \theta)$ to an empirical distribution $\pi(x, y)$ under
the log-loss. We denote by $x \in \mathcal{X}$ the observation vector and $y\in
\mathcal{Y}$ its associated label. Concretely, this stochastic optimization
problem aims to solve:
\begin{eqnarray}
\label{eq:objective}
\theta^*  & \in &
    \text{argmin}_{\theta}\
    \mathbb{E}_{(x,y) \sim \pi} \left[ - \log p(y \mid x, \theta) \right].
\end{eqnarray}

Defining the per-example loss as $\ell(x,y)$,
Stochastic Gradient Descent (SGD) performs the above minimization by
iteratively following the direction of steepest descent, given by the column
vector $\nabla = \mathbb{E}_\pi \left[ d\ell / d\theta \right]$.
Parameters are updated using the rule
$\theta^{(t+1)}\leftarrow \theta^{(t)} - \alpha^{(t)} \nabla^{(t)}$, where $\alpha$
is a learning rate.
An equivalent proximal form of gradient descent \cite{Combettes2009} reveals
the precise nature of $\alpha$:
\begin{eqnarray}
\label{eq:prox_sgd}
  \theta^{(t+1)} &=& \text{argmin}_{\theta}
  \left\{
    \left\langle \theta, \nabla \right\rangle +
    \frac{1}{2\alpha^{(t)}}\left\Vert \theta - \theta^{(t)} \right\Vert ^{2}_{2}
  \right\}
\end{eqnarray}
Namely, each iterate $\theta^{(t+1)}$ is the solution to an auxiliary optimization
problem, where $\alpha$ controls the distance between consecutive iterates,
using an $L_2$ distance.
In contrast, the natural gradient relies on the KL-divergence between iterates,
a more appropriate distance measure for probability distributions. Its metric is determined by
the Fisher Information matrix,
\begin{align}
    F_\theta = \mathbb{E}_{x \sim \pi} \left\{
         \mathbb{E}_{y \sim p(y\mid x, \theta)}
         \left[
            \left( \frac{\partial \log p}{\partial \theta} \right)
            \left( \frac{\partial \log p} {\partial \theta} \right)^T
         \right]
    \right\},
\end{align}
i.e. the covariance of the gradients of the model log-probabilities wrt. its
parameters. The natural gradient direction is then obtained as
$\nabla_N = F_\theta^{-1} \nabla$.
See \cite{Pascanu-natural-arxiv2013,Ollivier13} for a recent overview of the topic.

\subsection{Fisher Information Matrix for MLPs}
\label{sec:mlp_fisher}

We start by deriving the precise form of the Fisher for a canonical multi-layer perceptron (MLP)
composed of $L$ layers.
We consider the following deep network for binary classification, though
our approach generalizes to an arbitrary number of output classes.
\begin{eqnarray}
    \label{eq:mlp}
    p(y=1 \mid x) \equiv h_L &=& f_L(W_L h_{L-1} + b_{L}) \\
    \cdots \nonumber \\
    h_{1} &=& f_{1}\left(W_{1} x + b_{1}\right)  \nonumber
\end{eqnarray}

The parameters of the MLP, denoted
$\theta = \{W_1, b_1, \cdots, W_L, b_L\}$,
are the weights $W_i \in \mathbb{R}^{N_i \times N_{i-1}}$
connecting layers $i$ and $i-1$, and the biases $b_i \in \mathbb{R}^{N_i}$.
$f_i$ is an element-wise non-linear function.

Let us define $\delta_i$ to be the backpropagated gradient through the $i$-th
non-linearity.  We ignore the off block-diagonal components of the Fisher
matrix and focus on the block $F_{W_i}$, corresponding to interactions
between parameters of layer $i$.
This block takes the form:
\begin{align}
    \nonumber
    F_{W_i} &=  \mathbb{E}_{\substack{x\sim\pi \\ y\sim p}} \left[
        \matvec\left(\delta_i h_{i-1}^T\right)
        \matvec\left(\delta_i h_{i-t}^T\right)^T \right],
\end{align}
where $\matvec(X)$ is the vectorization function yielding a column vector
from the {\it rows} of matrix $X$.

Assuming that $\delta_i$ and activations $h_{i-1}$ are independent random
variables, we can write:
\begin{align}
    \label{eq:FIM_MLP}
    F_{W_i}(km, ln) & \approx
    \mathbb{E}_{\substack{x\sim\pi \\ y\sim p}} \left[ \delta_i(k) \delta_i (l) \right]
    \mathbb{E}_{\pi} \left[ h_{i-1}(m)  h_{i-1}(n) \right],
\end{align}
where $X(i,j)$ is the element at row $i$ and column $j$ of matrix $X$ and
$x(i)$ is the $i$-th element of vector $x$.
$F_{W_i}(km,ln)$ is the entry in the Fisher capturing interactions
between parameters $W_i(k,m)$ and $W_j(l,n)$.
Our hypothesis, verified experimentally in Sec.~\ref{sec:conditioning}, is that
we can greatly improve conditioning of the Fisher by enforcing that
$\mathbb{E}_{\pi} \left[ h_i h_i^T \right] = I$, for all layers of the network,
despite ignoring possible correlations in the $\delta$'s and off block
diagonal terms of the Fisher.

\section{Projected Natural Gradient Descent}
\label{sec:natmlp}

\begin{figure}
  \begin{tabular}{cc}
  \subfloat
  {
    \label{fig:natnn}
    \includegraphics[scale=0.7]{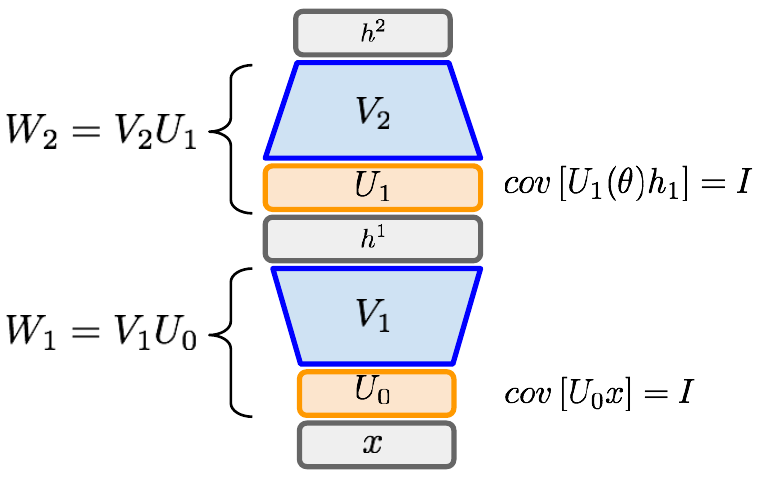}
  } &
  \hspace{1cm}
  \subfloat
  {
    \label{fig:apndg}
    \raisebox{.5cm}{\includegraphics[scale=0.35]{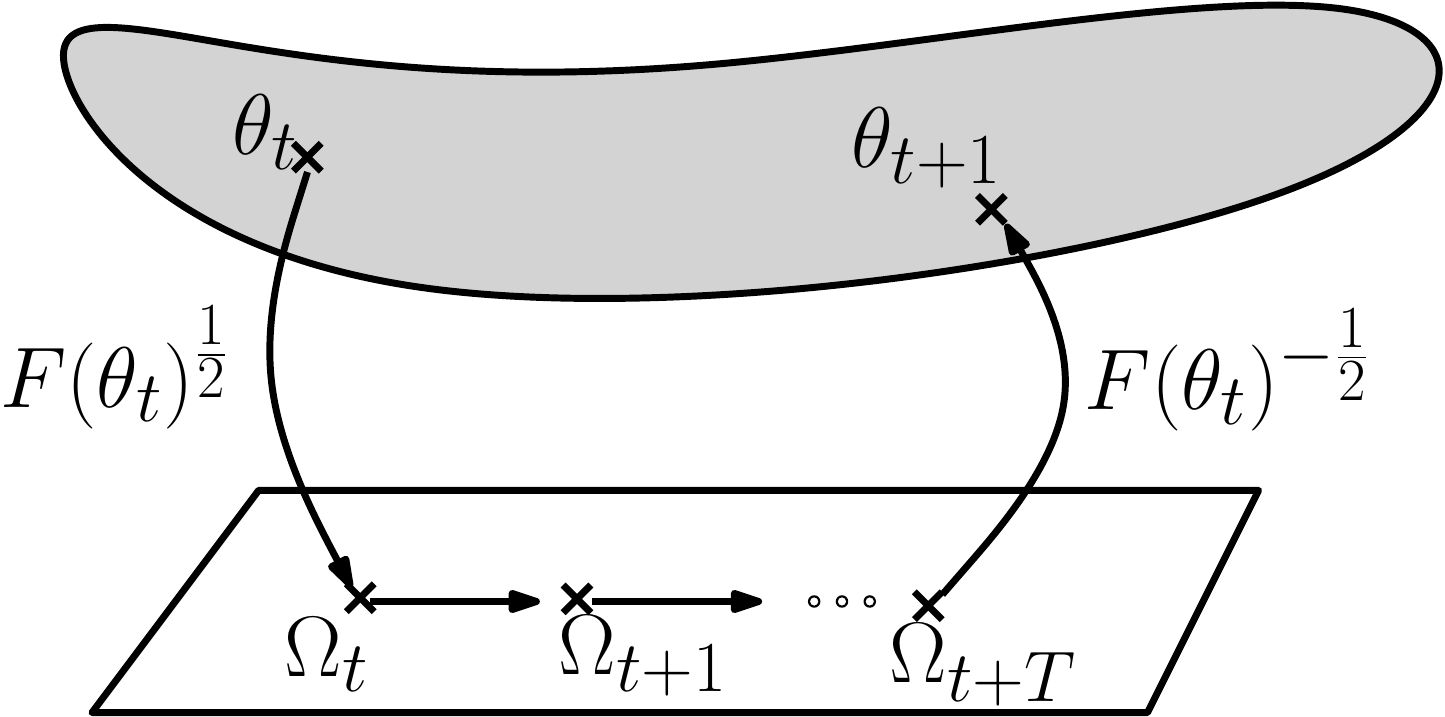}}
  }
  \end{tabular}
\caption{(a) A 2-layer natural neural network. (b) Illustration of the projections involved in PRONG.}
\end{figure}

This section introduces Whitened Neural Networks (WNN), which perform
approximate whitening of their internal hidden representations.
We begin by presenting a novel whitened neural layer, with the
assumption that the network statistics
$\mu_i(\theta) = \mathbb{E}[h_i]$ and
$\Sigma_i(\theta) = \mathbb{E}[h_i h_i^T]$
are fixed. We then show how these layers can be adapted to efficiently track
population statistics over the course of training. The resulting learning
algorithm is referred to as Projected Natural Gradient Descent (PRONG).  We
highlight an interesting connection between PRONG and Mirror Descent in
Section~\ref{sec:mirror}.

\subsection{A Whitened Neural Layer}

The building block of WNN is the following neural layer,
\begin{eqnarray}
    \label{eq:mlp2}
    h_{i} &=& f_{i}\left( V_i U_{i-1} \left( h_{i-1} - c_i \right) + d_{i} \right).
\end{eqnarray}

Compared to
Eq.~\ref{eq:mlp}, we have
introduced an explicit centering parameter $c_i=\mu_i$, which ensures that the input
to the dot product has zero mean in expectation. This is analogous to the centering
reparametrization for Deep Boltzmann Machines \cite{Montavon2013-centering}. The weight
matrix $U_{i-1} \in \mathbb{R}^{N_{i-1} \times N_{i-1}}$ is a per-layer
ZCA-whitening matrix whose rows are obtained from an eigen-decomposition of $\Sigma_{i-1}$:
\begin{align}
  \Sigma_i = \tilde{U}_{i}\cdot diag\left(\lambda_{i}\right)\cdot\tilde{U}_{i}^T
  \Longrightarrow
  U_{i} = diag\left(\lambda_{i} + \epsilon \right)^{-\frac{1}{2}}\cdot\tilde{U}_{i}^T.
\end{align}

The hyper-parameter $\epsilon$ is a regularization term controlling the maximal
multiplier on the learning rate, or equivalently the size of the trust region.
The parameters $V_i \in \mathbb{R}^{N_i \times N_{i-1}}$ and $d_{i} \in
\mathbb{R}^{N_i}$ are analogous to the canonical parameters of a neural network
as introduced in Eq.~\ref{eq:mlp}, though operate in the space of whitened
unit activations $U_i (h_i - c_i)$.
This layer can be stacked to form a deep neural network having $L$ layers, with
{\it model parameters} $\Omega=\{V_1, d_1, \cdots V_L, d_L\}$ and {\it whitening
coefficients}  $\Phi=\{U_0, c_0, \cdots, U_{L-1}, c_{L-1} \}$, as depicted in
Fig.~\ref{fig:natnn}.

Though the above layer might appear over-parametrized at first glance, we
crucially \emph{do not learn the whitening coefficients via loss minimization},
but
instead estimate them directly from the model statistics.
These coefficients are thus constants from the point of view of the optimizer
and simply serve to improve conditioning of the Fisher with respect to the
parameters $\Omega$, denoted $F_\Omega$. Indeed, using the same derivation that
led to Eq.~\ref{eq:FIM_MLP}, we can see that the block-diagonal terms of
$F_\Omega$ now involve terms $\mathbb{E}\left[ (U_i h_i) (U_i h_i)^T \right]$,
which equals identity by construction.

\subsection{Updating the Whitening Coefficients}

As the whitened model parameters $\Omega$ evolve during training, so do the
statistics $\mu_i$ and $\Sigma_i$. For our model to remain well conditioned,
the whitening coefficients must be updated at regular intervals, while taking
care not to interfere with the convergence properties of gradient descent. This
can be achieved by coupling updates to $\Phi$ with corresponding updates to
$\Omega$ such that the overall function implemented by the MLP remains
unchanged, e.g. by preserving the product $V_i U_{i-1}$ before and after each
update to the whitening coefficients (with an analoguous constraint on the
biases).

Unfortunately, while estimating the mean $\mu_i$ and $diag(\Sigma_i)$ could be
performed online over a mini-batch of samples as in the recent Batch
Normalization scheme \cite{batchnorm}, estimating the full covariance matrix
will undoubtedly require a larger number of samples.
While statistics could be accumulated online via an exponential moving average
as in RMSprop \cite{Rmsprop} or K-FAC \cite{martens2015}, the cost of the eigen-decomposition
required for computing the whitening matrix $U_i$ remains cubic in the layer size.

In the simplest instantiation of our method, we exploit the smoothness of gradient
descent by simply amortizing the cost of these operations over $T$ consecutive
updates. SGD updates in the whitened model will be closely aligned to NGD
immediately following the reparametrization. The quality of this approximation
will degrade over time, until the subsequent reparametrization.
The resulting algorithm is shown in the pseudo-code of
Algorithm~\ref{alg:pngd}. We can improve upon this basic amortization scheme by
including a diagonal scaling of $U_i$ based on the standard deviation
of layer $i$ activations, after each gradient update, thus mimicking the effect
of a diagonal natural gradient method. For this update to be valid, this
enhanced version of the method, denoted PRONG$^+$, scales the rows of $V_i$
accordingly so as to preserve the feed-forward computation of the network. 
This can be implemented by combining PRONG with batch normalization.

\setlength{\textfloatsep}{12pt}
\begin{algorithm}[t]
\caption{Projected Natural Gradient Descent}
\label{alg:pngd}
\begin{algorithmic}[1]
   \State {\bfseries Input:} training set $\mathcal{D}$, initial parameters $\theta$.
   \State {\bfseries Hyper-parameters:} reparam. frequency $T$, number of samples $N_s$, regularization term $\epsilon$.
   \State $U_i \leftarrow I; c_i \leftarrow 0; t \leftarrow 0$
   \Repeat
   \If{$mod(t, T) = 0$}
   \Comment{amortize cost of lines [6-11]}
     \For{all layers $i$}
     \State Compute canonical parameters $W_i = V_i U_{i-1}$; $b_i = d_i + W_i c_i$.
       \Comment{proj. $P^{-1}_\Phi(\Omega)$}
     \State Estimate $\mu_i$ and $\Sigma_i$, using $N_s$ samples from $\mathcal{D}$.
     \State Update $c_i$ from $\mu_i$ and $U_i$ from eigen decomp. of $\Sigma_i+\epsilon I$.
       \Comment{update $\Phi$}
     \State Update parameters $V_i \leftarrow W_i U_{i-1}^{-1}$; $c_i \leftarrow b_i - V_i c_i$.
       \Comment{proj. $P_\Phi(\theta)$}
     \EndFor
   \EndIf
   \State Perform SGD update wrt. $\Omega$ using samples from $\mathcal{D}$.
   \State $t \leftarrow t + 1$
   \Until{convergence}
\end{algorithmic}
\end{algorithm}

\subsection{Duality and Mirror Descent}
\label{sec:mirror}

There is an inherent duality between the parameters $\Omega$ of our whitened
neural layer and the parameters $\theta$ of a canonical model. Indeed,
there exist linear projections $P_\Phi(\theta)$ and $P_\Phi^{-1}(\Omega)$,
which map from canonical parameters $\theta$ to whitened parameters $\Omega$,
and vice-versa. $P_\phi(\theta)$ corresponds to line 10 of
Algorithm~\ref{alg:pngd}, while $P^{-1}_\Phi(\Omega)$ corresponds to line 7.
This duality between $\theta$ and $\Omega$ reveals a close connection between
PRONG and Mirror Descent \cite{Beck03}.

Mirror Descent (MD) is an online learning algorithm which generalizes the proximal
form of gradient descent to the class of Bregman divergences
$B_\psi(q, p)$, where $q, p \in \Gamma$ and $\psi: \Gamma \rightarrow \mathbb{R}$
is a strictly convex and differentiable function.
Replacing the L$_2$ distance by $B_\psi$, mirror descent solves the proximal problem
of Eq.~\ref{eq:prox_sgd} by applying first-order updates in a dual space and then
projecting back onto the primal space.
Defining
$\Omega = \nabla_\theta \psi (\theta)$ and
$\theta = \nabla_\Omega^* \psi (\Omega)$,
with $\psi^*$ the complex conjugate of $\psi$, the mirror descent updates are
given by:
\begin{eqnarray}
  \label{eq:md1}
  \Omega^{(t+1)} &=& \nabla_{\theta} \psi\left(\theta^{(t)}\right) - \alpha^{(t)} \nabla_\theta \\
  \label{eq:md2}
  \theta^{(t+1)} &=& \nabla_{\Omega} \psi^{*} \left(\Omega^{(t+1)}\right)
\end{eqnarray}
It is well known \cite{Thomas13,Raskutti13} that the natural gradient is a special case of
MD, where the distance generating function
\footnote{As the Fisher and thus $\psi_\theta$ depend on the parameters
$\theta^{(t)}$, these should be indexed with a time superscript, which we drop
for clarity.}
is chosen to be
$\psi(\theta) = \frac{1}{2} \theta^T F \theta$.

The mirror updates are somewhat unintuitive however. Why is the gradient
$\nabla_\theta$ applied to the dual space if it has been computed in the space
of parameters $\theta$ ? This is where PRONG relates to MD.
It is trivial to show that using the function
$\tilde{\psi}(\theta) = \frac{1}{2} \theta^T \sqrt{F} \theta$,
instead of the previously defined $\psi(\theta)$,
enables us to directly update the dual parameters using $\nabla_\Omega$,
the gradient computed directly in the dual space.
Indeed, the resulting updates can be shown to implement the natural gradient
and are thus equivalent to the updates of Eq.~\ref{eq:md2} with the appropriate
choice of $\psi(\theta)$:
\begin{eqnarray}
  \tilde{\Omega}^{(t+1)}
    &=& \nabla_{\theta} \tilde{\psi}\left(\theta^{(t)}\right) - \alpha^{(t)} \nabla_\Omega
    = F^{\frac{1}{2}} \theta^{(t)} - \alpha^{(t)} \mathbb{E}_\pi \left[ \frac{d\ell}{d\theta} F^{-\frac{1}{2}} \right] \nonumber \\
  \tilde{\theta}^{(t+1)}
    &=& \nabla_{\Omega} \tilde{\psi}^{*} \left(\tilde{\Omega}^{(t+1)}\right)
     = \theta^{(t)} - \alpha^{(t)} F^{-1} \mathbb{E}_\pi \left[ \frac{d\ell}{d\theta} \right] 
\end{eqnarray}

The operators $\tilde{\nabla}\psi$ and $\tilde{\nabla}\psi^*$ correspond to the
projections $P_\Phi(\theta)$ and $P^{-1}_\Phi(\Omega)$ used by PRONG to map from
the canonical neural parameters $\theta$ to those of the whitened layers
$\Omega$.  As illustrated in Fig.~\ref{fig:apndg}, the advantage of this whitened form
of MD is that one may amortize the cost of the projections over several
updates, as gradients can be computed directly in the dual parameter space.

\subsection{Related Work}

\begin{figure}[t!]
  \begin{tabular}{ccc}
  \subfloat[]
  {
    \label{fig:whitening_before}
    \raisebox{.5cm}{\includegraphics[scale=0.35]{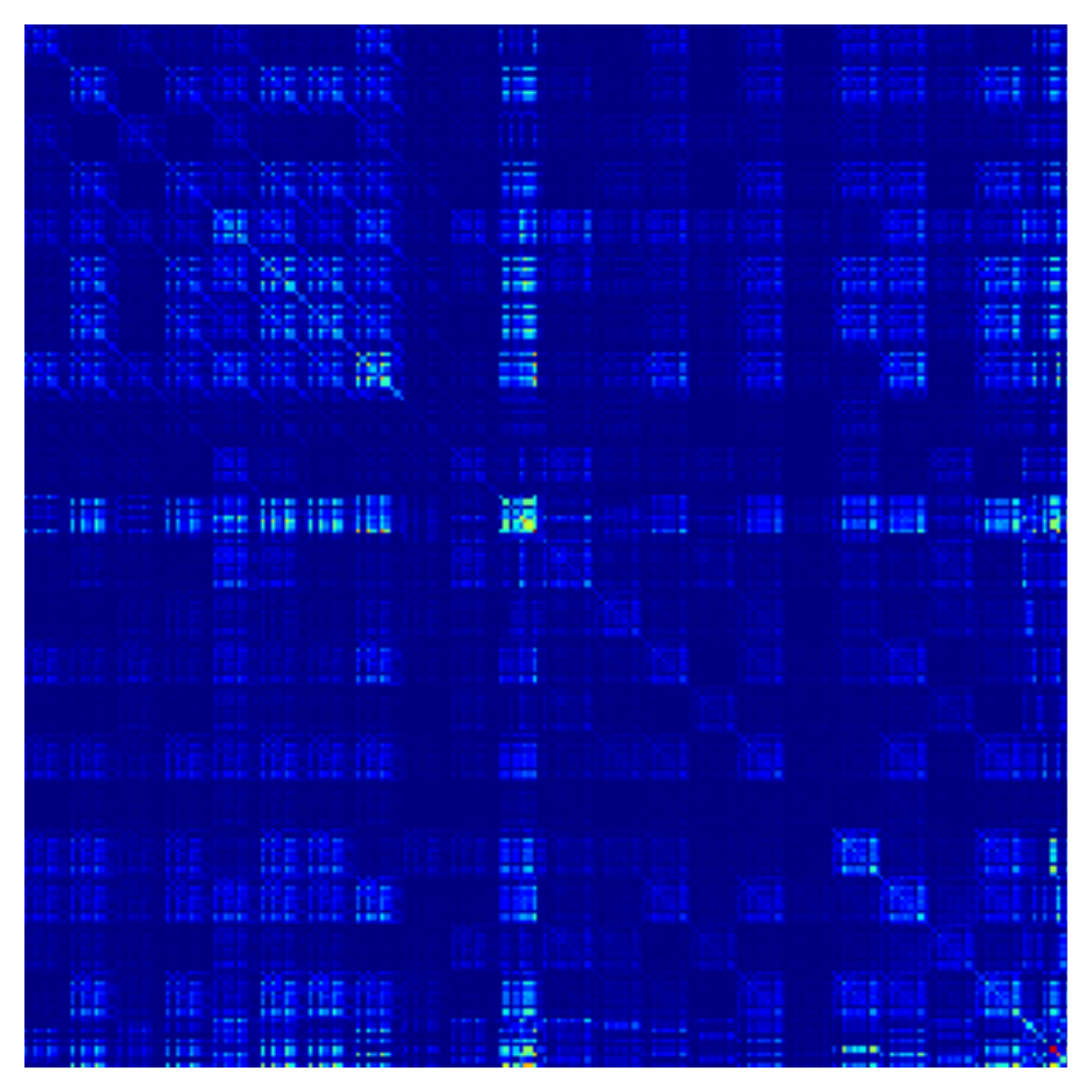}}
  } &
  \subfloat[]
  {
    \label{fig:whitening_after}
    \raisebox{.5cm}{\includegraphics[scale=0.35]{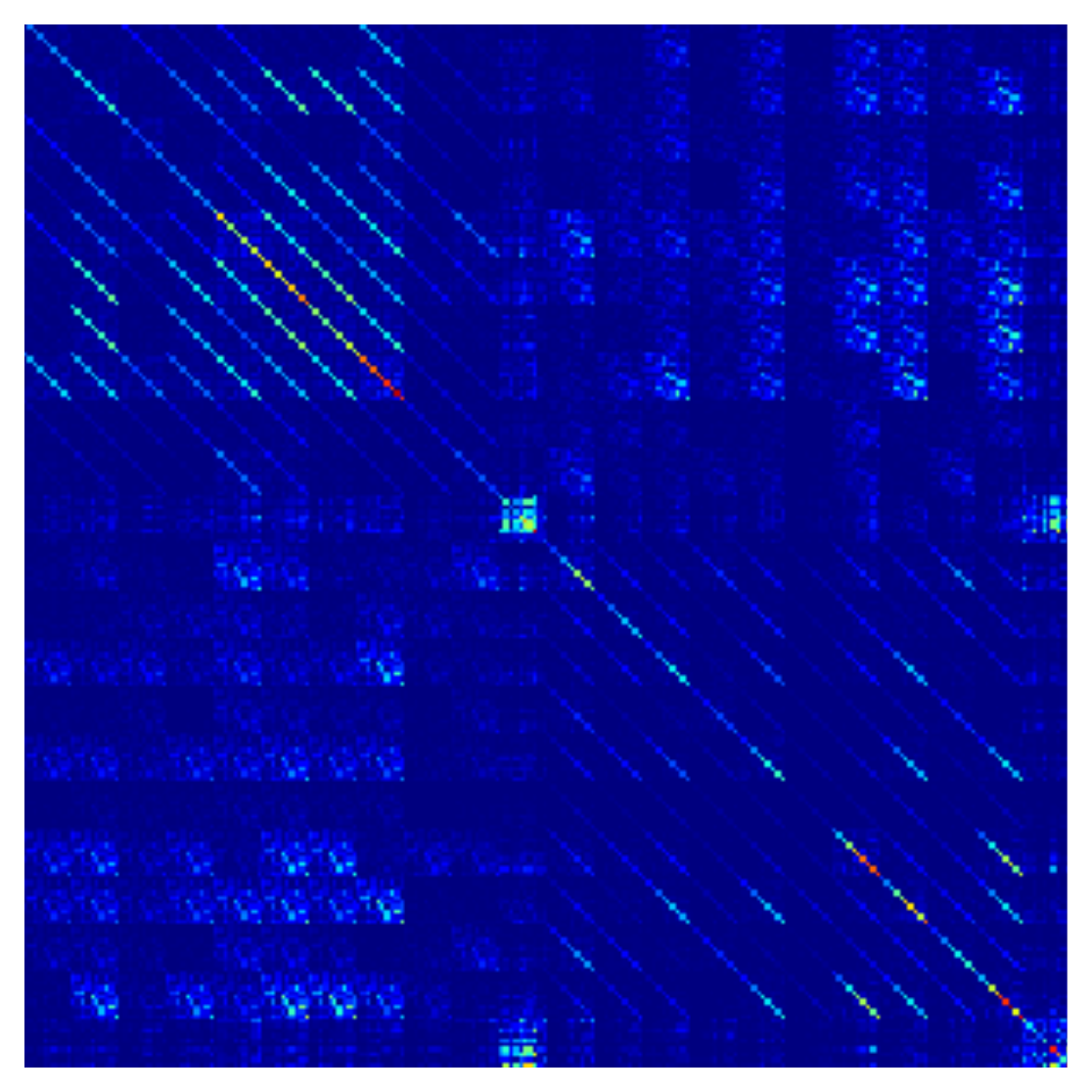}}
  } &
  \subfloat[]
  {
    \label{fig:conditional_number}
    \raisebox{.4cm}{\includegraphics[scale=0.35]{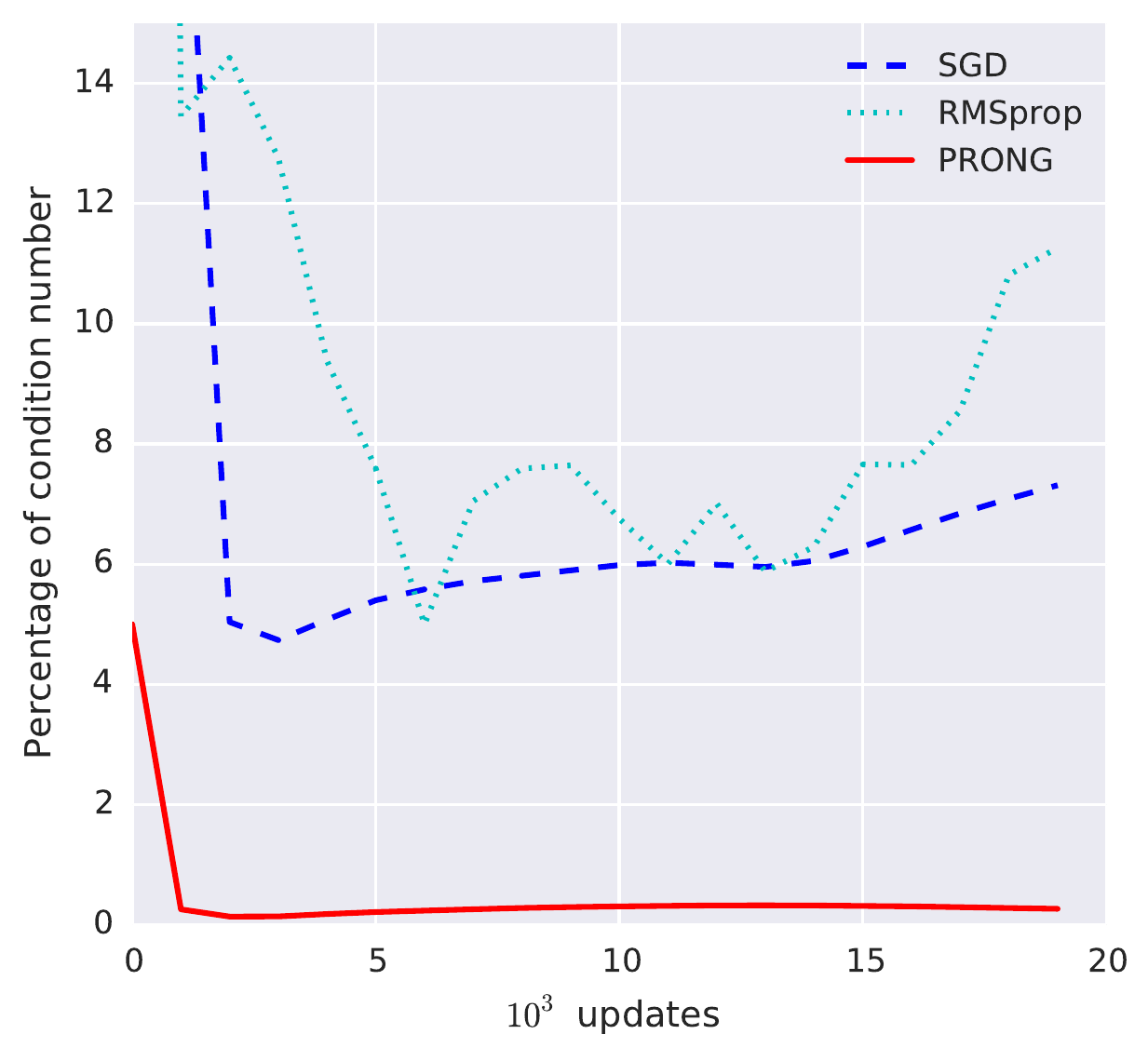}}
  }
  \end{tabular}
\caption{Fisher matrix for a small MLP (a) before and (b) after the first
  reparametrization. Best viewed in colour.
(c) Condition number of the FIM during training, relative to the initial conditioning.}
\label{fig:whitening_stats}
\end{figure}

This work extends the recent contributions of \cite{Raiko-2012-small} in
formalizing many commonly used heuristics for training MLPs: the
importance of zero-mean activations and gradients
\cite{LeCun+98backprop,Scraudolph98}, as well as the importance of normalized
variances in the forward and backward passes
\cite{LeCun+98backprop,Scraudolph98,GlorotAISTATS2010}.
More recently, \citet{Vatanen13} extended their previous work
\cite{Raiko-2012-small} by introducing a multiplicative constant $\gamma_i$ to
the centered non-linearity. In contrast, we introduce a full whitening matrix
$U_i$ and focus on whitening the feedforward network activations, instead of
normalizing a geometric mean over units and gradient variances.

The recently introduced batch normalization (BN) scheme \cite{batchnorm}
quite closely resembles a \emph{diagonal} version of PRONG, the main difference being
that BN normalizes the variance of activations {\it before} the non-linearity,
as opposed to normalizing the latent activations by looking at the full covariance.
Furthermore, BN
implements normalization by modifying the feed-forward computations thus
requiring the method to backpropagate through the normalization operator.
A diagonal version of PRONG also bares an interesting resemblance to RMSprop
\cite{Rmsprop,Duchi10}, in that both normalization terms involve the square root of the
FIM. An important distinction however is that PRONG applies this update in the
whitened parameter space, thus preserving the natural gradient interpretation.

K-FAC \cite{martens2015} is also closely related to PRONG and was developed
concurrently to our method. In one of its
implementations, it targets the same block diagonal as PRONG while also
exploiting the low rank structure of these blocks for efficiency,
reminiscent of TONGA\cite{LeRoux07}. Their method however operates online via
low-rank updates to each block, similar to the preconditioning used in the
Kaldi speech recognition toolkit \cite{Povey14}. This is in contrast to our
approach based on amortization.
They also consider the covariance of the backpropagated gradients $\delta_i$,
while PRONG only looks at the covariance of activations $h_i$. K-FAC further
proposes a tri-diagonal variant which decorrelates gradients across neighboring
layers, though resulting in a more complex algorithm.

A similar algorithm to PRONG was later found in \cite{Dickstein12}, where it
appeared simply as a thought experiment, but with no amortization or recourse
for efficiently computing $F$.

\section{Experiments}
\label{sec:exp}

\begin{figure}[t!]
  \hspace{-.8cm}
  \begin{tabular}{cccc}
  \subfloat[]
  {
    \label{fig:deepauto_covdiagreg}
    \includegraphics[height=0.457\linewidth]{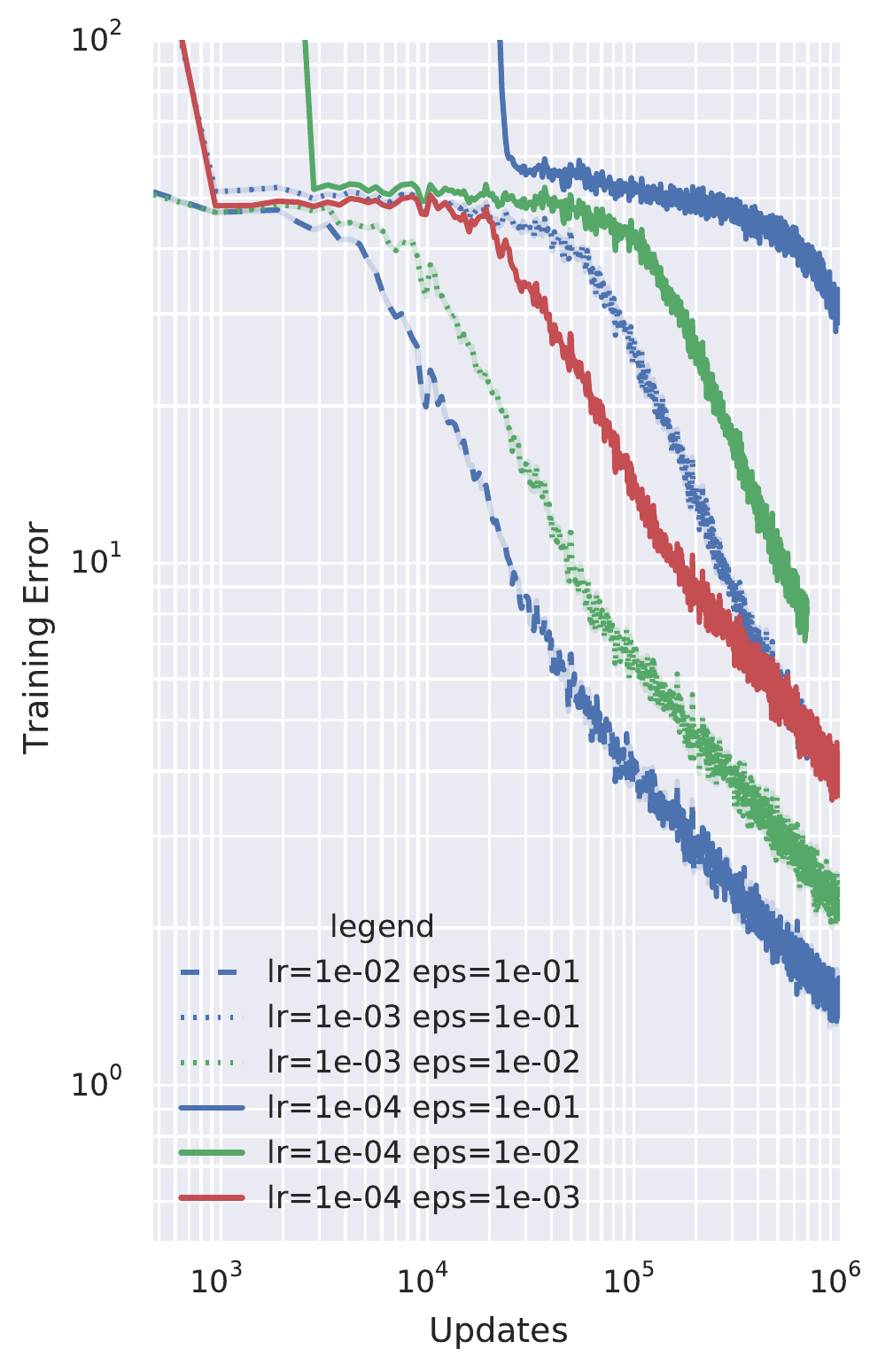}
  } &
  \hspace{-.8cm}
  \subfloat[]
  {
    \label{fig:deepauto_reparam}
    \includegraphics[height=0.45\linewidth]{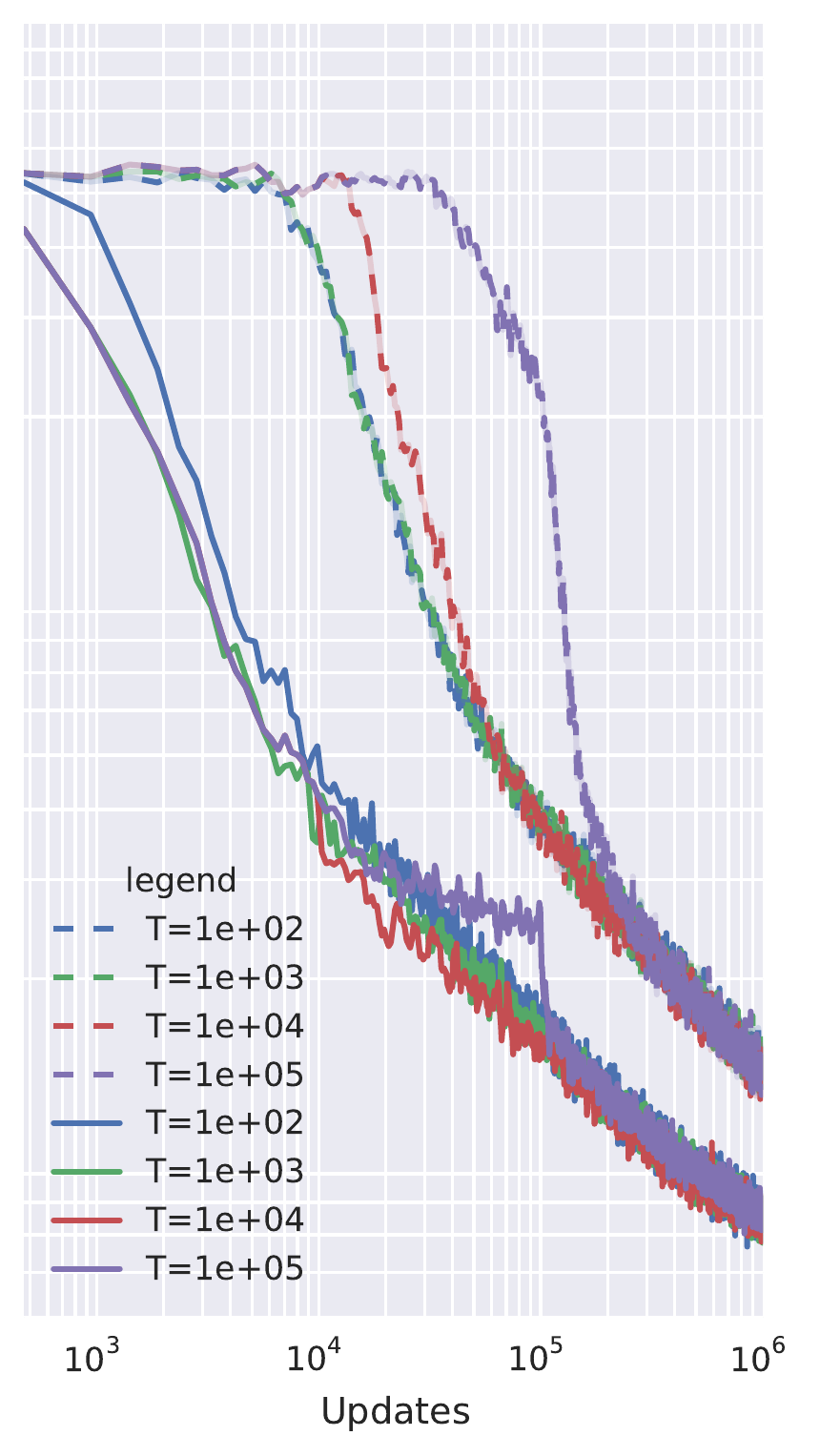}
  } &
  \hspace{-.8cm}
  \subfloat[]
  {
    \label{fig:deepauto_mnist_step}
    \includegraphics[height=0.45\linewidth]{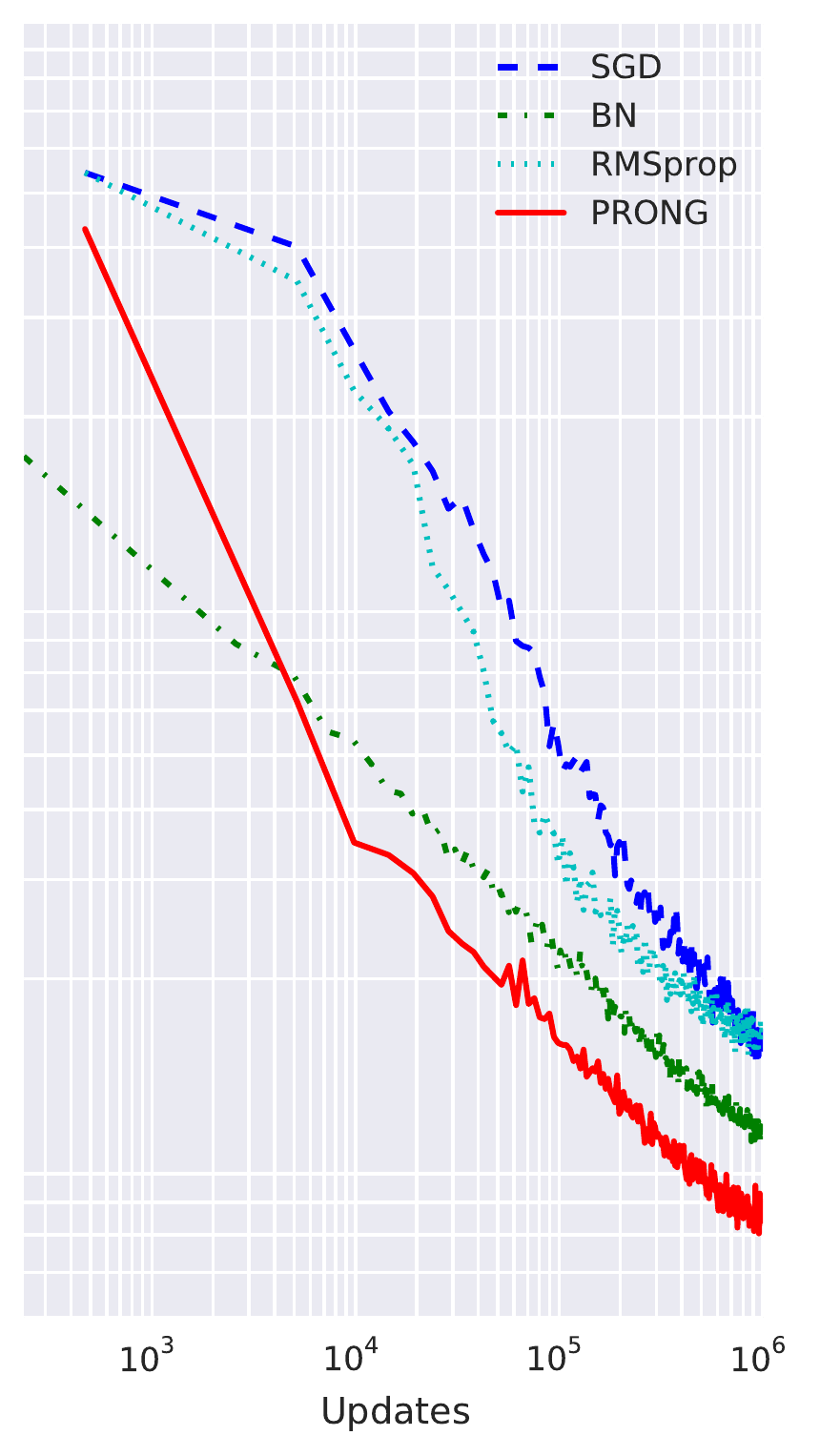}
  } &
  \hspace{-.8cm}
  \subfloat[]
  {
    \label{fig:deepauto_mnist_time}
    \includegraphics[height=0.45\linewidth]{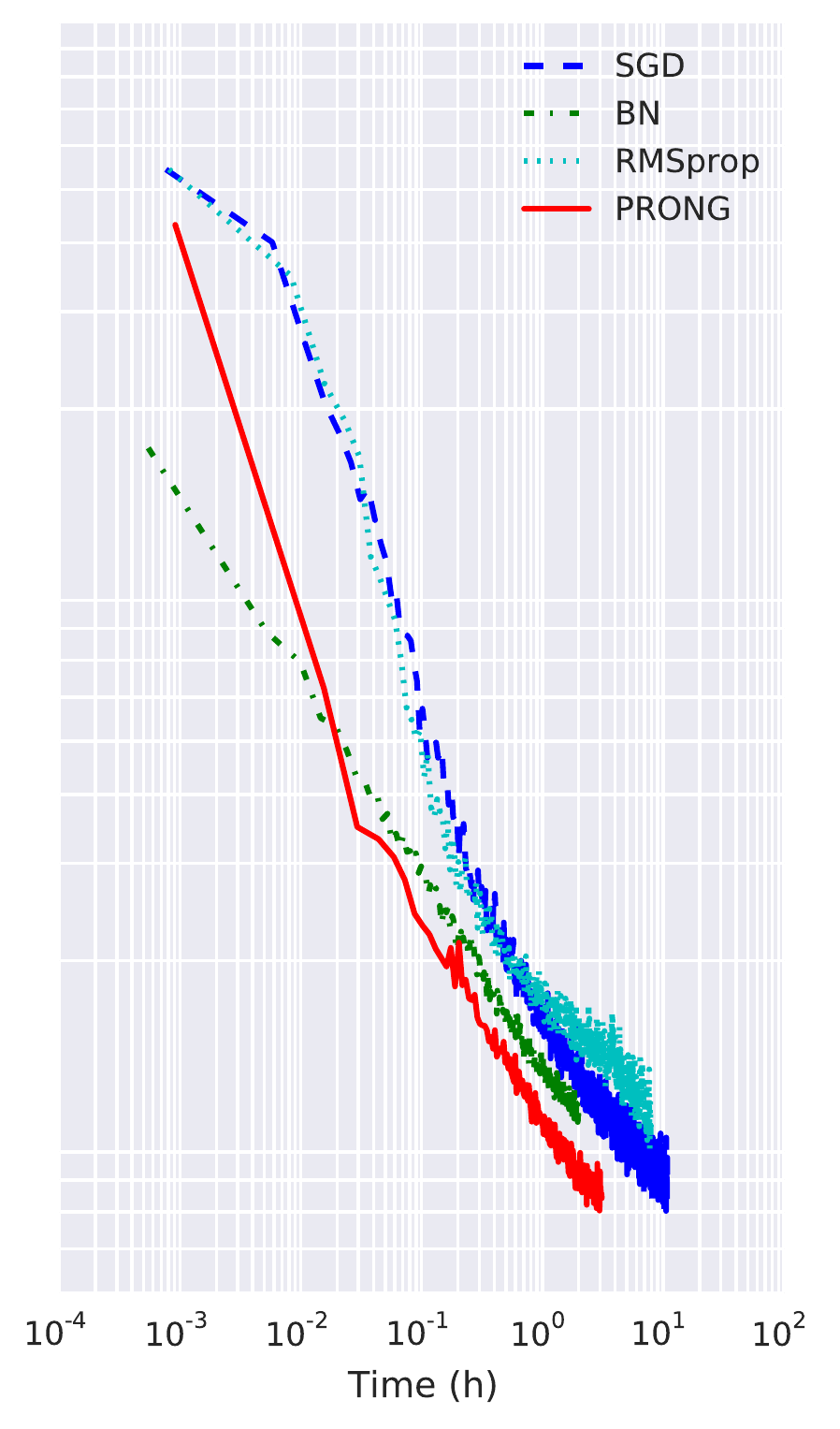}
  }
  \end{tabular}
\caption{Optimizing a deep auto-encoder on MNIST.
(a) Impact of eigenvalue regularization term $\epsilon$.
(b) Impact of amortization period $T$ showing that initialization with the whitening reparametrization is important for achieving faster learning and better error rate.
(c) Training error vs number of updates.
(d) Training error vs cpu-time. Plots (c) and (d) show that PRONG achieves better error rate both in number of updates and wall clock time. }
\label{fig:deepauto_mnist}
\end{figure}

We begin with a set of diagnostic experiments which highlight the effectiveness
of our method at improving conditioning. We also illustrate the impact of the
hyper-parameters $T$ and $\epsilon$, controlling the frequency of the
reparametrization and the size of the trust region.
Section~\ref{sec:unsup} evaluates PRONG on unsupervised learning problems, where
models are both deep and fully connected. Section~\ref{sec:sup} then moves onto
large convolutional models for image classification.

\subsection{Introspective Experiments}
\label{sec:conditioning}

\textbf{Conditioning.}
To provide a better understanding of the approximation made by PRONG, we train a small
3-layer MLP with tanh non-linearities, on a downsampled version of MNIST (10x10) \cite{Lecun98}.
The model size was chosen in order for the full Fisher to be tractable.
Fig.~\ref{fig:whitening_stats}(a-b) shows the FIM of the middle hidden layers before and after whitening the model activations (we took the absolute
value of the entries to improve visibility).
Fig.~\ref{fig:whitening_stats}c
depicts the evolution of the condition number of the FIM during training,
measured as a percentage of its initial value (before the first whitening
reparametrization in the case of PRONG). We present such curves for SGD, RMSprop and PRONG.
The results
clearly show that the reparametrization performed by PRONG improves conditioning
(reduction of more than 95\%). These observations confirm our initial
assumption, namely that we can improve conditioning of the block diagonal Fisher
by whitening activations alone.

\textbf{Sensitivity of Hyper-Parameters.}
Figures~\ref{fig:deepauto_covdiagreg}- \ref{fig:deepauto_reparam} highlight the
effect of the eigenvalue regularization
term $\epsilon$ and the reparametrization interval $T$. The experiments were performed on the best performing
auto-encoder of Section~\ref{sec:unsup} on the MNIST dataset.
Figures~\ref{fig:deepauto_covdiagreg}- \ref{fig:deepauto_reparam} plot the reconstruction error on the
training set for various values of $\epsilon$ and $T$.
As $\epsilon$ determines a maximum multiplier on the learning rate, learning becomes
extremely sensitive when this learning rate is high\footnote{Unstable combinations
of learning rates and $\epsilon$ are omitted for clarity.}.
For smaller step sizes however, lowering $\epsilon$ can yield significant speedups
often converging faster than simply using a larger learning rate.
This confirms the importance of the manifold curvature for optimization (lower $\epsilon$
allows for different directions to be scaled drastically different according to their
corresponding curvature).
Fig~\ref{fig:deepauto_reparam} compares the impact of $T$ for models having
a proper whitened initialization (solid lines), to models being initialized
with a standard ``fan-in'' initialization (dashed lines) \cite{LeCun+98backprop}. These results are quite surprising
in showing the effectiveness of the whitening reparametrization as a simple
initialization scheme. That being said, performance can degrade due to ill
conditioning when $T$ becomes excessively large ($T=10^5$).

\subsection{Unsupervised Learning}
\label{sec:unsup}

Following \citet{martens2010hessian}, we compare PRONG on the
task of minimizing reconstruction error of an 8-layer auto-encoder on the MNIST
dataset.  The encoder is composed of 4 densely connected sigmoidal layers, with
a number of hidden units per layer in $\{1\text{k},500,250,30\}$, and a
symmetric (untied) decoder. Hyper-parameters were selected by grid search,
based on training error, with the following grid specifications:
training batch size in $\{32, 64, 128, 256\}$, learning rates in
$\{10^{-1}, 10^{-2}, 10^{-3}\}$ and momentum term in $\{0, 0.9\}$. For RMSprop,
we further tuned the moving average coefficient in $\{0.99, 0.999\}$ and the
regularization term controlling the maximum scaling factor in $\{0.1, 0.01\}$.
For PRONG, we fixed the natural reparametrization to $T=10^3$, using $N_s=100$
samples (i.e. they were not optimized for wallclock time).
Reconstruction error with respect to updates and wallclock time are shown in
Fig.~\ref{fig:deepauto_mnist} (c,d).

We can see that PRONG significantly outperforms the baseline methods, by up to
an order of magnitude in number of updates. With respect to wallclock, our
method significantly outperforms the baselines in terms of time taken to reach a certain error threshold, despite the fact that the runtime per epoch for
PRONG was 3.2x that of SGD, compared to batch normalization (2.3x SGD) and RMSprop (9x SGD). Note that these timing numbers reflect performance under the optimal choice of hyper-parameters, which in the case of batch normalization yielded a batch size of $256$, compared to $128$ for all other methods.
Further breaking down the performance, $34\%$ of the runtime of PRONG was spent
performing the whitening reparametrization, compared to $4\%$ for estimating
the per layer means and covariances. This confirms that amortization is paramount
to the success of our method.\footnote{We note that our implementation of the whitening operations is not optimized, as it does not take advantage of GPU acceleration, as opposed to the neural network computations. Therefore, runtime of our method is expected to improve as we move the eigen-decompositions to GPU.}

\subsection{Supervised Learning}
\label{sec:sup}

The next set of experiments addresses the problem of training deep supervised
convolutional networks for object recognition.
Following \cite{batchnorm}, we perform whitening across feature
maps only: that is we treat pixels in a given feature map as independent
samples. This allows us to implement the whitened neural layer as a sequence of
two convolutions, where the first is by a 1x1 whitening filter. PRONG is
compared to SGD, RMSprop and batch normalization, with each algorithm being
accelerated via momentum.
Results are presented on both CIFAR-10 \cite{Krizhevsky09} and the ImageNet Challenge
(ILSVRC12) datasets \cite{ILSVRC15}.
In both cases, learning rates were decreased using a ``waterfall'' annealing
schedule, which divided the learning rate by $10$ when the validation error failed
to improve after a set number of evaluations.
\footnote{
  On CIFAR-10, validation error was estimated every $10^3$
  updates and the learning rate decreased by a factor of $10$ if the validation
  error failed to improve by $1\%$ over $4$ consecutive evaluations.
  For ImageNet, we employed a more aggressive schedule which required that the
  validation error improves by $1\%$ after each epoch.
}

\begin{figure}[t!]
  \begin{tabular}{cccc}
  \hspace{-.8cm}
  \subfloat[]
  {
    \label{fig:cifar10_train_steps}
    \includegraphics[width=0.27\linewidth]{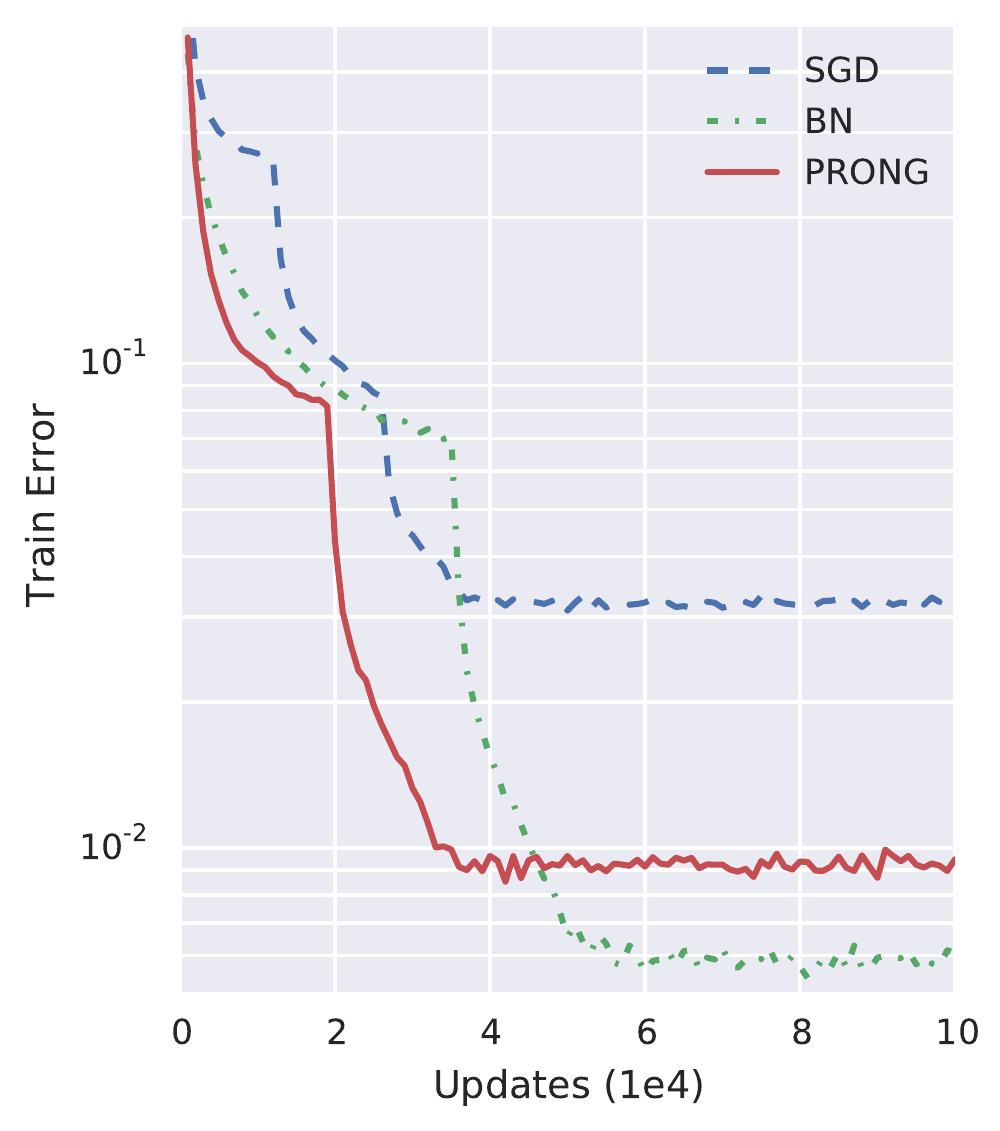}
  } &
  \hspace{-.8cm}
  \subfloat[]
  {
    \label{fig:cifar10_test_time}
    \includegraphics[width=0.27\linewidth]{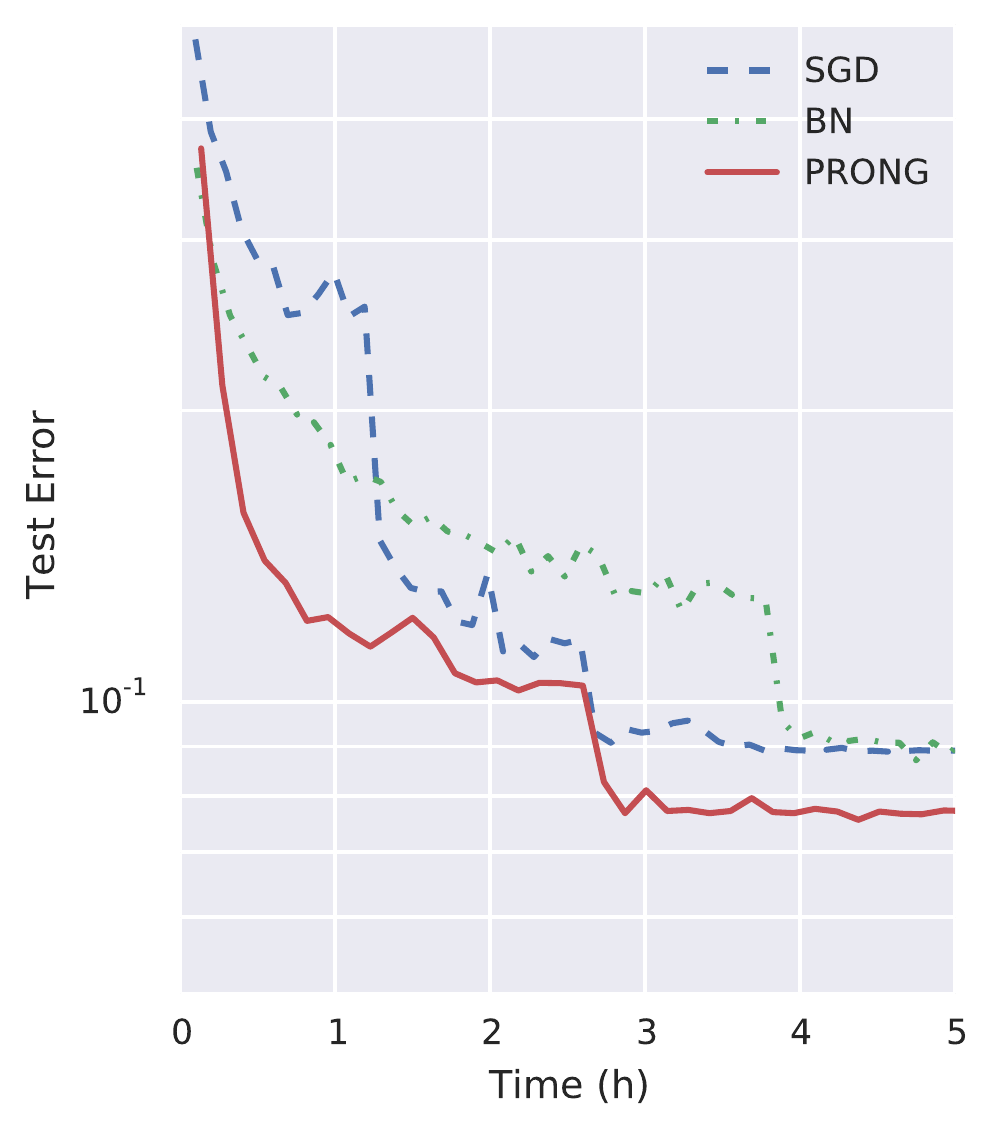}
  } &
  \hspace{-.8cm}
  \subfloat[]
  {
    \label{fig:imnet_train_steps}
    \includegraphics[width=0.27\linewidth]{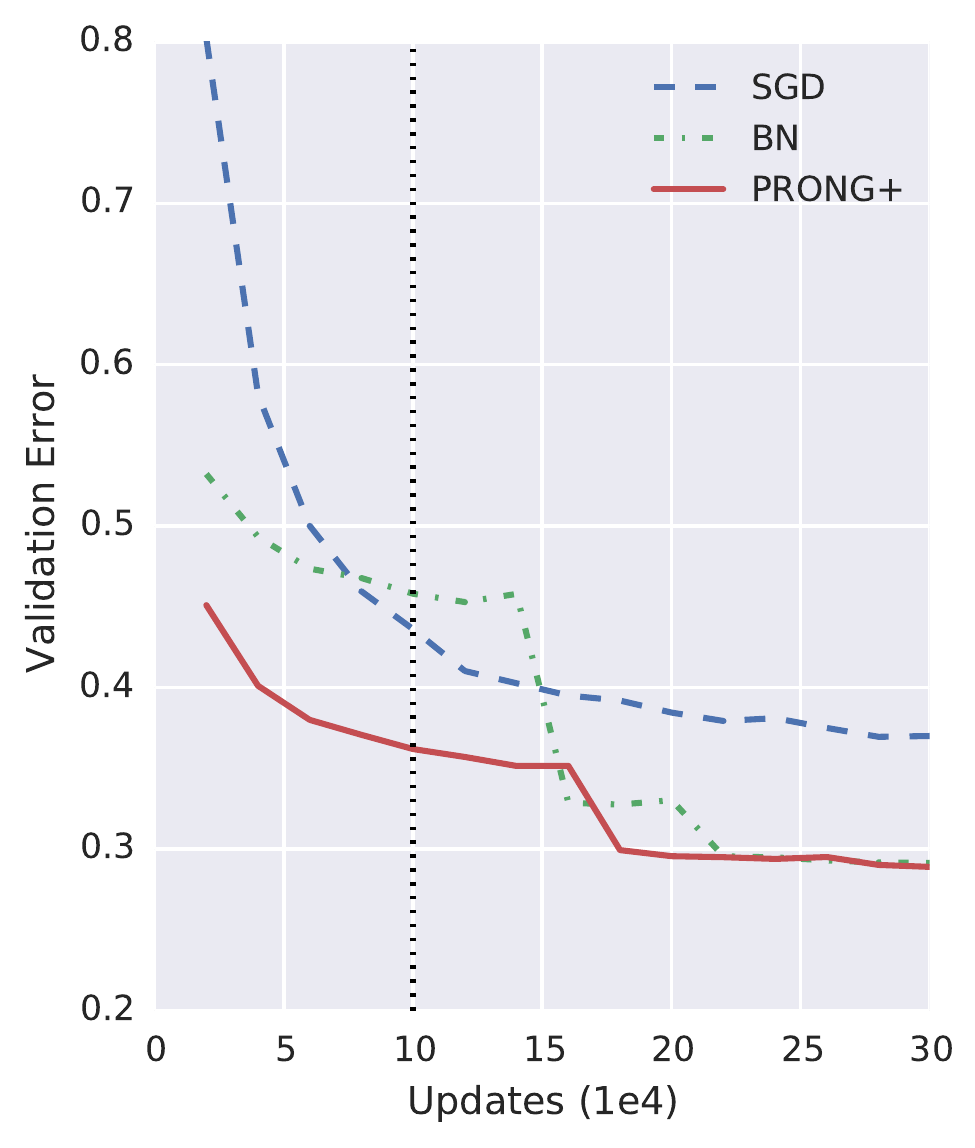}
  } &
  \hspace{-.8cm}
  \subfloat[]
  {
    \label{fig:imnet_test_time}
    \includegraphics[width=0.27\linewidth]{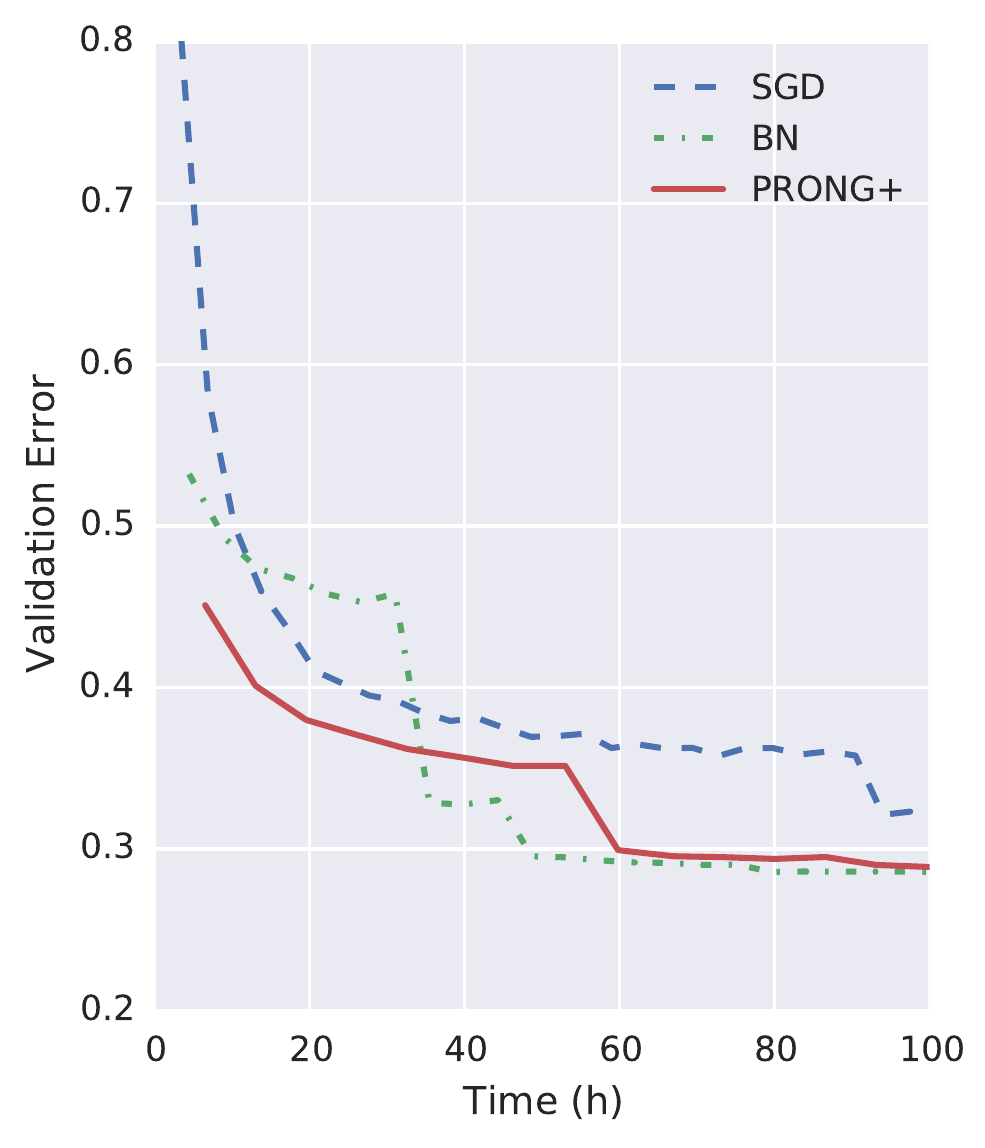}
  }
  \end{tabular}
  \caption{Classification error on CIFAR-10 (a-b) and ImageNet (c-d). On CIFAR-10, PRONG achieves better test error and converges faster. On ImageNet, PRONG$^+$ achieves comparable validation error while maintaining a faster covergence rate.}
\label{fig:cifar_imnet_results}
\end{figure}

\subsubsection{CIFAR-10}

The model used for our CIFAR experiments consists of 8 convolutional layers,
having $3\times 3$ receptive fields. $2\times 2$ spatial max-pooling was applied
between stacks of two convolutional layers, with the exception of the last convolutional
layer which computes the class scores and is followed by global max-pooling and
soft-max non-linearity.
This particular choice of architecture was inspired by the VGG model
\cite{Simonyan15} and held fixed across all experiments. The number of filters
per layer is as follows: $64, 64, 128, 128, 256, 256, 512, 10$. The model was
trained on $24\times 24$ random crops with random horizontal reflections.
Model selection was performed on a held-out validation set of $5$k examples.
Results are shown in Fig.~\ref{fig:cifar_imnet_results}.

With respect to training error, PRONG and batch normalization seem to offer
similar speedups compared to SGD with momentum. Our hypothesis is that the
benefits of PRONG are more pronounced for densely connected networks, where the
number of units per layer is typically larger than the number of maps used
in convolutional networks. Interestingly, PRONG generalized better, achieving
$7.32\%$ test error vs. $8.22\%$ for batch normalization. This could reflect
the findings of \cite{Pascanu-natural-arxiv2013}, which showed how NGD
can leverage unlabeled data for better generalization: the ``unlabeled'' data
here comes from the extra perturbations in the training set when estimating the
whitening matrices.

\subsubsection{ImageNet Challenge Dataset}

Our final set of experiments aims to show the scalability of our method: we thus
apply our natural gradient algorithm to the large-scale ILSVRC12 dataset (1.3M images labelled into 1000 categories)
using the Inception architecture \cite{batchnorm}. In order to scale to problems
of this size, we parallelized our training loop so as to split the processing of
a single minibatch (of size $256$) across multiple GPUs. Note that PRONG can scale well in this
setting, as the estimation of the mean and covariance parameters of each layer
is also embarassingly parallel.
Eight GPUs were used for computing gradients and estimating
model statistics, though the eigen decomposition required for whitening was
itself not parallelized in the current implementation.

For all optimization algorithms,
we considered initial learning rates in $\{10^{-1}, 10^{-2}, 10^{-3}\}$ and used a value
of $0.9$ as the momentum coefficient.
For PRONG
we tested reparametrization periods $T\in \{10,10^2,10^3,10^4\}$, while
typically using $N_s = 0.1 T$. Eigenvalues were regularized by
adding a small constant $\epsilon \in \{1, 10^{-1}, 10^{-2}, 10^{-3}\}$ before
scaling the eigenvectors
\footnote{The grid was not searched exhaustively as the cost would have been
prohibitive. As our main focus is optimization, regularization
consisted of a simple $L_2$ weight decay parameter of $10^{-4}$, with no Dropout
\cite{Srivastava14}.}.
Given the difficulty of the task, we employed the enhanced PRONG$^+$
version of the algorithm, as simple periodic whitening of the model proved to
be unstable.\footnote{This instability may have been compounded by momentum,
which was initially not reset after each model reparametrization when using
standard PRONG.}

Figure~\ref{fig:cifar_imnet_results} (c-d) shows that batch normalisation and
PRONG$^+$ converge to approximately the same top-1 validation error ($28.6\%$ vs $28.9\%$
respectively) for similar cpu-time. In comparison, SGD achieved a
validation error of $32.1\%$. PRONG$^+$ however exhibits much faster
convergence initially: after $10^5$ updates it obtains around
$36\%$ error compared to $46\%$ for BN alone. We stress that the ImageNet
results are somewhat preliminary.
While our top-1 error is higher than reported in \cite{batchnorm}
($25.2\%$), we used a much less extensive data augmentation pipeline.
We are only beginning to explore what natural gradient methods may achieve on
these large scale optimization problems and are encouraged by these initial
findings.

\section{Discussion}

We began this paper by asking whether convergence speed could be improved by
simple model reparametrizations, driven by the structure of the Fisher
matrix.
From a theoretical and experimental perspective, we have shown that Whitened
Neural Networks can achieve this via a simple, scalable and efficient
whitening reparametrization. They are however one of several possible instantiations of the concept of Natural
Neural Networks. In a previous incarnation of the idea, we exploited a similar
reparametrization to include whitening of backpropagated
gradients\footnote{The weight matrix can be parametrized as $W_i = R_i^T
V_i U_{i-1}$, with $R_i$ the whitening matrix for $\delta_i$.}.
We favor the simpler approach presented in this paper, as we
generally found the alternative less stable with deep networks. Ensuring
zero-mean gradients also required the use of skip-connections, with tedious
book-keeping to offset the reparametrization of centered non-linearities
\cite{Raiko-2012-small}.

Maintaining whitened activations may also offer additional benefits
from the point of view of model compression and
generalization. By virtue of whitening, the projection $U_i h_i$ forms an
ordered representation, having least and most significant bits. The sharp
roll-off in the eigenspectrum of $\Sigma_i$ may explain why deep
networks are ammenable to compression \cite{NIPS2014_5484}. Similarly, one could envision spectral
versions of Dropout \cite{Srivastava14} where the dropout probability is a
function of the eigenvalues.
Alternative ways of orthogonalizing the representation at each layer should
also be explored, via alternate decompositions of $\Sigma_i$, or perhaps
by exploiting the connection between linear auto-encoders and PCA. We
also plan on pursuing the connection with Mirror Descent and further bridging
the gap between deep learning and methods from online convex optimization.

\subsubsection*{Acknowledgments}
We are extremely grateful to Shakir Mohamed for invaluable discussions and
feedback in the preparation of this manuscript. We also thank Philip Thomas,
Volodymyr Mnih, Raia Hadsell, Sergey Ioffe and Shane Legg for feedback on the paper.

\small{\bibliography{paper}}
\bibliographystyle{plainnat}

\end{document}